\documentclass{article} 
\usepackage{iclr2016_conference,times}
\usepackage{hyperref}
\usepackage{url}

\title{Super-resolution with deep convolutional sufficient statistics}

\author{Joan Bruna\\
 Department of Statistics\\
University of California, Berkeley\\
\texttt{joan.bruna@berkeley.edu}
\And
Pablo Sprechmann \\
Courant Institute\\
New York University\\
\texttt{pablo@cims.nyu.edu}  \\
\And
 Yann LeCun \\
Facebook, Inc., \&\\
Courant Institute\\
New York University\\
\texttt{yann@cims.nyu.edu}  \\
}

\usepackage{amsmath,graphicx}
\usepackage{amsmath}
\usepackage{amssymb}
\usepackage{graphicx}
\usepackage{caption}
\usepackage{subcaption}
\usepackage{wrapfig}
\usepackage[titletoc,title]{appendix}

\usepackage{bm}

\newcommand{\R}{\mathbb{R}}

\iclrfinalcopy 

\begin{document}

\maketitle

\begin{abstract}
Inverse problems in image and audio, and super-resolution in particular, can be seen as high-dimensional structured prediction problems, where the goal is to characterize the conditional distribution of a high-resolution output given its low-resolution corrupted observation. When the scaling ratio is small, point estimates achieve impressive performance, but soon they suffer from the regression-to-the-mean problem, result of their inability to capture the multi-modality of this conditional distribution. 
Modeling high-dimensional image and audio distributions is a hard task, requiring both the ability to model complex geometrical structures and textured regions. 
In this paper, we propose to use as conditional model a Gibbs distribution, where its sufficient statistics are given by deep Convolutional Neural Networks (CNN). The features computed by the network are stable to local deformation, and have reduced variance when the input is a stationary texture. These properties imply that the resulting sufficient statistics minimize the uncertainty of the target signals given the degraded observations, while being highly informative.
The filters of the CNN are initialized by multiscale complex wavelets, and then we propose an algorithm to fine-tune them by estimating the gradient of the conditional log-likelihood, which bears some similarities with Generative Adversarial Networks. We evaluate experimentally the proposed framework in the image super-resolution task, but the
approach is general and could be used in other challenging ill-posed problems such as audio bandwidth extension.

%
%
%
\end{abstract}

\section{Introduction}
\label{intro}


Single-image super resolution aims to construct a high-resolution image from a single low-resolution input. 
Traditionally, these type of inverse problems in image and signal processing are approached by constructing  models with appropriate priors to regularize the signal estimation. 
Classic Tychonov regularization considers Sobolev spaces as a means to characterize the regularity of the desired solutions.
While numerically very efficient, such priors are not appropriate in most applications, since 
typical signals are not globally smooth.

More recently, data-driven approaches have lead to very successful results in the context of inverse problems, as
training data can be used as a means to adjust the prior to the empirical distribution.
Most methods in the literature fall into two main categories: non-parametric and
parametric. The former aim to obtain the co-occurrence prior between the high-resolution and low-resolution local structures from an external
training database. The high-resolution instance is produced by copying small patches from the most similar examples found in the training set (\cite{freeman2002example}).
Parametric models with sparse regularization have enjoyed great success in this problem (\cite{elad2006image,yang2008image}), thanks to their capacity to capture local 
regularity with efficient convex optimization methods (\cite{mairal2009online}). 
Recent works have shown that better results can be obtained 
when the models are tuned on recovery performance rather
than on data fitting (\cite{yang2012coupled,mairal2012task}). However, the non-explicit form of the estimator leads to a more 
difficult bi-level optimization problem (\cite{mairal2012task}). This difficulty can be mitigated by learning approximate inference mechanisms that take advantage of the 
specifics of the data on which they are applied (\cite{LecunNN,pami15}).

With the mindset of maximizing recovery performance, several works have proposed to replace the inference step by a generic neural network architecture having enough capacity to perform non-linear regression (\cite{dong2014learning,cui2014deep}). This approach closely relates with the auto-encoders paradigm extensively studied in the representation
learning literature. Auto-encoders learn mid level-features using an information-preservation criterion, maximizing
recoverability of the training signals from the extracted features (\cite{autoencoders}). 

In the context of super resolution, the systems are trained as to minimize a measure of fitness between the ground truth signal and its reconstruction from the distorted observation. There is an intrinsic limitation in this approach: the mapping being approximated is highly unstable (or even multi-valued).
The biggest challenge to overcome is the design of an objective
function that encourages the system to discover meaningful regularities and produce sensible estimations.
The most popular objective is the squared Euclidean distance in the signal domain, despite the fact it is not correlated to good perceptual quality, as it is not stable to small deformations and uncertainty leads to linear blurring, a well known effect commonly referred as \emph{regression to the mean}.

In order to model the multi-modality intrinsic in many inverse problems, it is thus necessary to replace point estimates by  inferential models. A popular approach is to use variational inference over a certain mixture graphical model.
If $x \in \R^M$ are the low-resolution observations and $y \in \R^N$ (commonly with $N\gg M$) are the target, high-resolution samples, 
we might attempt to model the conditional distribution $p(y ~|~x)$ via a collection of hidden variables, in order to account for the multi-modality. Variational autoencoders (\cite{kingma2013auto}) and other recent generative models showed that one can encode relatively complex geometrical properties by mapping separable latent random variables with a deep neural network (\cite{gsn,adversarial1, adversarial2, gregor2015draw, sohl2015deep, rezende2015variational}) .

However, a question remains open: can these models scale and  account for highly non-Gaussian, stationary processes that form  textured regions in images and sounds, without incurring in an explosion on the number of hidden variables? In this paper, we take a different approach. Instead of considering a mixture model of the form $p(y | x) = \int p(y | x, h) p(h|x) dh$, 
we propose to learn a non-linear representation of the target signal $\Psi(y)$ that expresses this multi-modal distribution in terms of a Gibbs density:
\begin{equation}
\label{eq.modo}
p(y | x) \propto \exp( - \| \Phi(x) - \Psi(y) \|^2),
\end{equation}
where $\Phi:\R^N\rightarrow \R^P$ and  $\Psi:\R^M\rightarrow \R^P$ are non-linear mappings that take, respectively, observations and targets to a common high-dimensional space of dimension, $P$. In other words, we propose to learn a collection of non-linear sufficient statistics $\Psi(y)$ that minimize the uncertainty of $y$ given $x$ while being highly informative. In order for (\ref{eq.modo}) to be an efficient conditional model, the features $\Psi(y)$ must therefore reduce the uninformative variability while preserving discriminative information.

We argue that a good parametric model for such sufficient statistics are given by Convolutional Neural Networks (CNNs) (\cite{Lecun98}), since they incorporate two important aspects in their architecture. On the one hand, they provide stability to small geometric deformations thanks to the rectifier and pooling units. On the other hand, when the input is a locally stationary process or texture, they provide features with smaller variance. These properties can be proved for a certain class of CNNs where filters are given by multi-scale wavelets (\cite{scatt}), but remain valid over a certain region of the CNN parameter space. Learning non-linear convolutional features that capture natural image statistics has also been considered in the context of unsupervised learning (\cite{henaff2014local}), as well as Gibbs models whose sufficient statistics are given by deep convolutional features \cite{Dai2015ICLR}.
Our approach also relates to the method proposed by \cite{gatys2015texture} for texture synthesis. The authors show that the texture representations obtained across the layers of a CNN increasingly
capture the statistical properties of natural images, producing impressive texture synthesis results. The recent work by \cite{lamb2016discriminative} proposed to use representations derived from CNNs trained on discriminative tasks to regularize the objective function of the variational autoencoder. 
Unlike the work presented in this paper, these methods use representations obtained from CNNs trained in a purely discriminative fashion, without further adaptation. 


In summary, our main contributions are:
\begin{itemize}
\item We develop a framework for solving inverse problems in a suitable non-linear representation space.
\item We propose an algorithm to fine-tune the collection of sufficient statistics on a conditionally generative model.
\item We demonstrate the validity of the approach on a challenging ill-posed problem: image super-resolution. 
\end{itemize}

The rest of the paper is organized as follows. 
In Section~\ref{sec:setup}, we present the set-up of the super-resolution problem. Then, we describe the proposed approach Section~\ref{model}.
In Section~\ref{experiments} we analyze the proposed approach experimentally using examples of image super-resolution.  We conclude the paper and discuss future research directions in Section~\ref{conclusions}.

\section{Problem Set-Up}
\label{sec:setup}

We consider the task of estimating a high-dimensional vector $y \in \mathbb{R}^N$ given 
observations $x = U(y) \in \mathbb{R}^M$. In all problems of interest, the operator $U$ is not invertible. 
In the case of image or audio super-resolution, $U$ performs a downsampling. In absence of training data, inverse problems are approached as regularized signal recovery problems,
 leading to programs of the form 
$$\hat{y} = \arg\min_y \frac{1}{2} \| U(y) - x \|^2 + \lambda \mathcal{R}(y)~,$$
where $\mathcal{R}$ captures the structure one wishes to enforce amongst the infinity of possible solutions matching the observations. Possible regularizations include total variation norms of the form $\mathcal{R}(y) = \| \nabla y \|_1$ or 
sparsity in predefined signal decompositions, such as wavelets. 

%
Training data can be incorporated into the problem as a means to adjust the prior to the empirical distribution.
In this section we describe the setting in which the estimation is treated as pure regression problem, leading to a point estimation problem of the form 
\begin{equation}
\label{dnnsr}
\min_{\Theta} \sum_i \| \Phi( x_i, \Theta) - y_i \|^2~,
\end{equation}
where $(x_i, y_i)$ are training examples and $\Phi(x, \Theta)$ is a CNN parametrized by $\Theta$. 
Model (\ref{dnnsr}) can be interpreted probabilistically by associating the mean squared loss with the negative log-likelihood of a Gaussian model:
\begin{equation}
\label{bla1}
p(y ~|~x) = \mathcal{N} ( \Phi(x, \Theta), {\bf Id} )~.
\end{equation}
In that case, one is only interested in the point estimate $\hat{y}=\Phi(x,\Theta)$, which corresponds  to the Maximum Likelihood Estimate (MLE) of (\ref{bla1}). 

Although appealing for its simplicity, the previous formulation has a limitation, that comes from the instability induced by the inverting the operator $U\!$. 
If $y_1$ is a signal with sharp structures, and $y_2$ is such that $y_2(u) = y_1(u -\tau(u))$ is a small deformation of $y_1$, 
then one has $\| U y_1 - U y_2 \| \approx 0$ if the warping field $\tau$ is sufficiently small, although $\| y_1 - y_2 \| $ might be of the order of $\|y_1\|$.\footnote{More precisely, one can verify that the Euclidean distance is proportional to $\xi \| \tau \|$ where $\xi$ is the central frequency of $y_1$. }  
In other words, the high-frequency information of $y$ is intrinsically unstable to geometric deformations under the standard Euclidean metric. It results 
that a point estimate optimizing the Euclidean distance as in (\ref{dnnsr}) will necessarily discard the high frequency responsible for the large difference $\| y_1 - y_2\|$.

Moreover, if $y$ is a stationary process, such as those modeling auditory or image textures, the point estimate will converge to the conditional expectation $\mathbb{E}( y ~|~ Uy=x)$, which in general does not contain the same amount of high-frequency information than typical realizations of $y$. 
In both previous examples, the point estimate exhibits the so-called ``regression-to-the-mean" phenomena, which is accentuated as the scaling ratio of the operator $U$ increases. Our goal in the next section is to develop an alternative framework that combines the simplicity of point estimates while capturing high-frequency information.

\section{Inference Model}
\label{model}

In this section we develop our energy model based on a collection of stable sufficient statistics. 
We begin by motivating the model, then we present examples of predefined statistics, 
and next we propose an algorithm to fine-tune those statistics to the data. 

\subsection{From Point Estimates to stable Inference}
\label{sec.stabel_inf}


Similarly as in the work by \cite{adversarial2}, we model the conditional distribution of high resolution samples given the low-resolution input through the residuals of a linear prediction. In other words, given a pair $(x, y)$, we define $\bar{x}=\bar{U}x$ to be the best linear predictor (in terms of Mean Squared Error (MSE)) of $y$ given $x$, and consider the task of modeling the residual $r= y- \bar{U}(x)$, which carries all the high-frequency information. 

In this work, we consider a conditional model defined by a Gibbs energy of the form 
\begin{equation}
\label{bla2}
\| \Phi(x) - \Psi(r) \|^2~,
\end{equation}
where $\Phi$ and $\Psi$ are a pair of generic CNNs. 
As with every energy-based model, (\ref{bla2}) admits a probabilistic interpretation, by considering the corresponding
Gibbs distribution
\begin{equation}
\label{bla3}
p( r|~x) = \exp \left( -\| \Phi(x) - \Psi(r) \|^2 - \log Z \right)~,
\end{equation}
where $Z$ is the partition function.
The model (\ref{bla2}), when viewed as a (conditionally) generative model of $r$, is thus characterized by a 
collection of sufficient statistics $\Psi(r)$. In particular, the invariance properties of $\Psi$ 
translate directly into iso-probability sets: if $r$, $r'$ are such that $\Psi(r)=\Psi(r')$, then $p(r ~|~x) = p(r'~|~x)$.

The model (\ref{bla2}) thus breaks the estimation task into two sub-problems: (i) approximate a collection 
of features $\Psi(r)$ using any parametric regression conditional on $x$, $\Phi(x)$, and (ii) sample 
a candidate $r'$ such that $\Psi(r')$ matches the resulting features. These two problems are connected by  
the trade-off caused by the amount of information contained in the sufficient statistics $\Psi$. If $\Psi$ is nearly invertible, then 
we put all the pressure on the approximation step, but the level sets of $\Psi$ are nearly singular, thus making the subsequent inference transparent. On the other hand,
if the features $\Psi(r)$ are too compressive (for instance, $\Psi(r) = \|r \|$), the approximation step is alleviated, but 
the inference of $r$ becomes powerless since the level sets of $\Psi$ will have too large dimensionality. 

It is thus necessary to find the right balance between stability and discriminability in the design of the sufficient statistics.
Motivated by the success on supervised learning in image and audio tasks, we consider signal representations, $\Psi$, given by complex CNNs. The deep convolutional architecture provides generic stability with respect to small geometric deformations, thanks to the rectification and average pooling layers. Moreover, the average pooling reduces the variance of locally stationary processes. At the same time, CNN features have the ability to achieve such local invariance while being highly informative (\cite{pooling_recovery, scatt}). 

The main question is thus how to determine $\Psi$. This problem has been studied extensively in the energy-based learning literature, see (\cite{lecun2006tutorial}) and references therein. Gibbs models with sufficient statistics given by deep representations have been previously
considered in the context of unsupervised learning in \cite{ngiam2011learning, Dai2015ICLR} and concurrently with our work in \cite{xiearxiv}. 
Before addressing the learning of $\Psi$, we discuss the setup where $\Psi$ is a predetermined CNN with specific properties.

Finally it is worth mentioning that the model can also be seen as the nonlinear equivalent of Canonical Correlation Analysis (CCA). 
Whereas CCA is concerned with finding linear transformations of both $x$ and $r$ that optimize the cross-correlation structure, thus maximizing predictability, our model considers non-linear transformations in both predictors and responses that will be maximally correlated.

%
%

\begin{figure}
\centering
\includegraphics[width=0.7\textwidth]{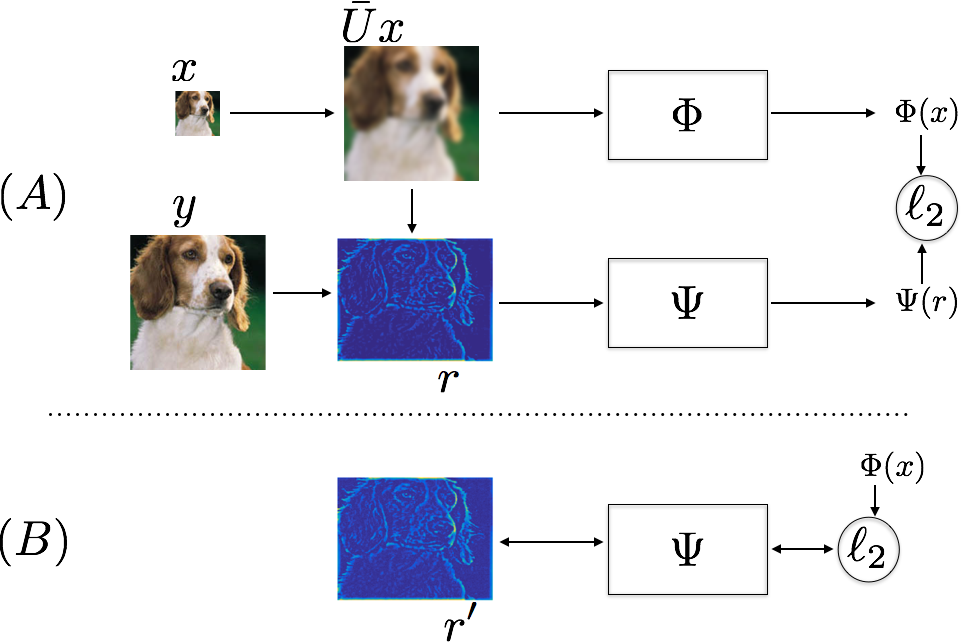}
\caption{Model overview. (A) For a given pair $(x,y)$, we first preprocess the data to obtain the high-frequency residuals $r$. The sufficient statistics $\Psi$ are computed for $r$, and these are approximated by a network $\Phi$. 
(B) The sampling procedure: Given $x$, we obtain a mode of the distribution $p(r~|~x)$ by solving $\min_r \| \Psi(r) - \Phi(x) \|^2$. }
\end{figure}

\subsection{Predefined CNN statistics}

\subsubsection{Scattering}
\label{sec:scatt}

Scattering networks (\cite{scatt, scatt_stephane}) are obtained by cascading complex wavelet decompositions with complex modulus. 
The wavelet decompositions consist of oriented complex band-pass filters that span uniformly the frequency plane. 
The resulting scattering coefficients are thus local averages of cascaded complex wavelet modulus coefficients $|\,|x \ast \psi_1| \ast \psi_2 | \ast \phi$ for different combination of scales and orientations.
Thanks to their particular multi-scale structure, scattering coefficients guarantee a number of stability properties 
in face of geometric deformations and stationary processes (\cite{scatt_stephane}).  
Moreover, scattering representations can be inverted under some conditions, which suggest that
a regression in the scattering metric is more tolerant to perceptually small high-frequency variations than the original 
Euclidean metric. 

Whereas scattering networks used in recognition applications typically contain only steerable filters (\cite{scatt}), in our setup it is worth pointing out that one can recover traditional total variation norms by properly adjusting the first wavelet decomposition. Indeed, if we consider an extra orientation at the finest scale of the form 
$$\psi_h(u) = \nabla_x(u) + i \nabla_y(u)~, $$
where $\nabla_x$ and $\nabla_y$ are respectively the discrete horizontal and vertical derivative operators, then the resulting first order scattering coefficient $ | x \ast \psi_h| \ast \phi = | \nabla x | \ast \phi $ is a local total variation norm. The model thus contains predictions of local total variation based on the low-resolution samples. Shrinking the resulting Total Variation (TV) prediction is an effective technique to regularize the estimation, similarly to wavelet shrinkage estimators, that generalizes the flat TV prior.

For a scattering decomposition with $J$ scales, $L$ orientations, and two layers of non-linearities, one has $O(J^2 L^2)$ scattering coefficients per patch of size $2^J$. For a given downsampling factor $\alpha>1$ (typically $\alpha=2,3,4$), the number of effective scattering coefficients visible in the low-resolution $x$ is $O( (J -\log(\alpha))^2 L^2)$, which gives a rough indication of how to adjust the scale and dimensionality of the scattering representation as a function of the downsampling factor. 

\subsubsection{Pre-trained discriminative networks}
\label{sec:vgg}

Recent findings in computer vision problems have shown that the representations learned for supervised classification problems can be readily transferred
 to others tasks (\cite{oquab2014learning}). Thus, another possibility is to transfer convolutional features learnt on a large supervised task.
As such an example, we use one of the networks proposed by \cite{vgg}. 
These network follow the standard CNN architecture with the particularity of using
very small ($3\times 3$) convolution filters, which enables the use of networks with significantly larger depth compared to prior approaches.
Specifically, we use the VGG-19 network, containing 16 convolutional layers,  5 max-pooling layers,  three fully connected layers and ReLU non-linearities.
We define the network $\Psi$ as a truncated VGG-19 network. 
Through testing on a small validation set, we found that best performance is achieved by keeping only the layer up to the forth pooling layer and 
replacing the max-pooling layers with average pooling ones, as the gradient flow is smother.
Note that keeping higher level representations entails higher computational cost. At testing time, this truncated network can be applied on images of arbitrary sizes.
In Section~\ref{sec.stabel_inf}, we model the conditional distribution of high resolution samples given the low-resolution input through the residuals of a linear prediction.
The main practical advantage of this choice is to reduce model complexity. When using pre-trained networks, however, this is not an option.
The VGG networks were trained with full images, thus it is not clear whether the representation obtained for residual images would be at all meaningful.
Hence, when using the VGG networks, $\Psi$ takes the high-resolution image $y$ as an input.

\subsection{Training predictive model}

We train our predictive model $\Phi(x)$ by directly minimizing (\ref{bla2}) with respect to the parameters of $\Phi$, which amounts to solving for a point estimate in the transformed feature space. 
Notice, however, that this is not equivalent to directly optimizing (\ref{bla3}) with respect to the parameters of $\Phi$. Indeed, contrary to the linear case, the partition function $Z$ depends on $x$. By optimizing the approximation error in (\ref{bla2}) we obtain a fast, suboptimal solution, which assumes that the approximation errors in the feature space are isotropic (i.e. the direction of the error does not influence the resulting likelihood). This assumption could be mitigated by replacing the Euclidean distance in (\ref{bla2}) with a Mahalanobis distance. At test-time, given a low-resolution query $x_0$, samples $r_0$ are produced by solving 
an optimization problem as explained in the next section.

\subsection{Sampling}
\label{sec:sampling}

An essential component of the model is a method to sample from a distribution of the form 
$p(r~|~x) \propto \exp( -\| \Phi(x) - \Psi(r) \|^2)$ when $\Psi$ is a non-linear, not necessarily invertible transformation. 

In particular, we will be interested in sampling a mode of this distribution (which in general will not be unique), which amounts to solving 
\begin{equation}
\label{optim}
\min_{r'} \| \Phi(x) - \Psi(r') \|^2~.
\end{equation}
Similarly as other works (\cite{simoncelli, henaff2014local}), we solve (\ref{optim}) using gradient descent, by initializing $r'= \bar{U}(x)$, where $\bar{U}$ is a Point Estimate (which can be linear or non-linear). From a computational perspective, this sampling procedure is significantly more expensive than evaluating the feed-forward baseline CNN. The gradient descent strategy requires running as many forward and backard passes of $\Psi$ as iterations. 
When using the pre-trained networks, (\ref{optim}) is solved directly over the high resolution image and the gradient descent optimization is initialized with the the best linear predictor of $y$ given $x$.

\subsection{Fine-tuning sufficient statistics}
\label{sec:fine}

As discussed previously, the design of good sufficient statistics requires striking the right balance between stability with respect to perturbations that are ``invisible" by the operator $U$, and discriminability, so that the iso-probability sets contain only relevant samples. We study in this section a model to adjust automatically this trade-off by approximating the gradient of the log-likelihood.

Given a conditional model of the form, 
\begin{equation}
p(r | x) = \exp( -|| \Phi(x) - \Psi(r) ||^2 - \log Z ) 
\label{eq.loss}
\end{equation}
we shall consider optimizing the negative log-likelihood in an alternate fashion with respect to $\Phi$ and $\Psi$. 
By taking logs and using the log-partition definition, we verify that 
\begin{eqnarray*}
\nabla_\Psi -\log p(r | x) &=& -\nabla \Psi(r)^T ( \Phi(x) - \Psi(r) ) + \mathbb{E}_{r' \sim p(r' | x)}  \nabla \Psi(r')^T ( \Phi(x) - \Psi(r') ) ~, \\
\nabla_\Phi -\log p(r | x) &=& \nabla \Phi(x)^T ( \Phi(x) - \Psi(r) ) - \nabla \Phi(x)^T \left(\Phi(x) - \mathbb{E}_{r' \sim p(r' | x)}   (\Psi(r')) \right) \\
&=& \nabla \Phi(x)^T \left( \mathbb{E}_{r' \sim p(r' | x)} (\Psi(r')) - \Psi(r)\right)~.
\end{eqnarray*}
An estimator of the gradients is obtained by sampling $p(r'|x)$ and replacing the expectation by sample averaging:
\begin{eqnarray*}
\widehat{\nabla_\Psi -\log p(r | x)} &=& -\nabla \Psi(r)^T ( \Phi(x) - \Psi(r) ) + \frac{1}{L} \sum_{r' \sim p(r' | x)}  \nabla \Psi(r')^T ( \Phi(x) - \Psi(r') )~,\\
\widehat{\nabla_\Phi -\log p(r | x)} &=& \nabla \Phi(x)^T \left( \frac{1}{L} \sum_{r' \sim p(r' | x)} \Psi(r') - \Psi(r) \right) ~,
\end{eqnarray*}
which is unbiased even for $L=1$.
Sampling from the exact conditional $p(r'|~x)$ requires an MCMC algorithm, such as Metropolis-Hastings, which would be computationally very expensive.  
In order to accelerate the fine-tuning process, we consider the following biased sampling procedure. Instead of sampling from the true Gibbs distribution, $p(r'|x)$, we obtain typical samples having large likelihood by solving (\ref{optim}) every time we require a new sample. This algorithm can be viewed as a co-area representation of $p(r'~|x)$: we first sample a high likelihood value, represented by a small approximation error $\| \Phi(x) - \Psi(r') \|^2$, and then we obtain a sample from the corresponding iso-probability set by randomly perturbing the initialization. 

Using gradient descent as a means to approximately sample from Gibbs distributions was proposed in \cite{mumford}, and later exploited in \cite{simoncelli} amongst other works. Although under some ergodicity assumptions one can show that the Gibbs distribution converges to the uniform measure on the a set of the form $\{ r \,; \Psi(r) = \psi_0 \}$, there are no guarantees that a gradient descent will produce samples from that uniform distribution.

The resulting model thus bears some resemblances with the Generative Adversarial Network framework introduced by \cite{adversarial1}, later extended by \cite{adversarial2}. The main difference is that in our case the discriminator and the generator networks are associated with a feature representation $\Psi$ and its inverse, respectively. In other words, the ``fake" samples are generated by inverting the discriminator in its current state.

%
%
%
%
%
%
%

%
%


\begin{figure}
        \begin{subfigure}[b]{0.24\textwidth}
\includegraphics[width=\textwidth]{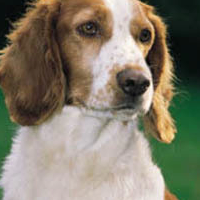}\\[-2ex]
        \caption{Original \newline Size: $200\times200$ pixels \newline}
\label{exp1_A}
        \end{subfigure}
                \begin{subfigure}[b]{0.24\textwidth}
        \includegraphics[width=\textwidth]{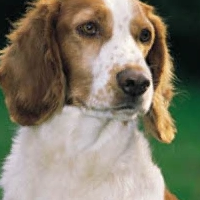}\\[-2ex]
        \caption{Scattering \newline PSNR: 37.54dB \newline}
                \end{subfigure}
        \hfill
        \begin{subfigure}[b]{0.24\textwidth}
\includegraphics[width=\textwidth]{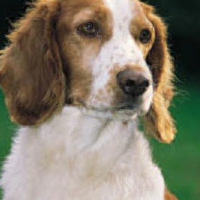}\\[-2ex]
\caption{VGG-19 4th level  \newline PSNR: 34.13dB  \newline}
        \end{subfigure}
        \hfill
        \begin{subfigure}[b]{0.24\textwidth}
\includegraphics[width=\textwidth]{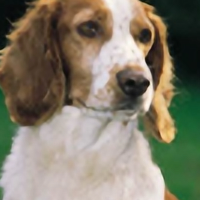}\\[-2ex]
\caption{Baseline \newline PSNR: 34.77dB \newline}
        \end{subfigure}\\[-4ex]

                \begin{subfigure}[b]{0.24\textwidth}
        \includegraphics[width=\textwidth]{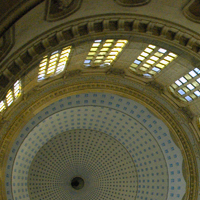}
        \caption{Original \newline Size: $200\times200$ pixels}
        \label{exp1_B}
                \end{subfigure}
                                \hfill
                                \begin{subfigure}[b]{0.24\textwidth}
                        \includegraphics[width=\textwidth]{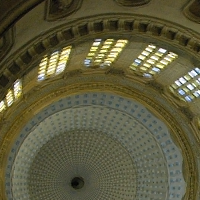}
                        \caption{Scattering \newline PSNR: 24.76dB}
        \label{exp1_F}
                                \end{subfigure}
                \hfill
                \begin{subfigure}[b]{0.24\textwidth}
        \includegraphics[width=\textwidth]{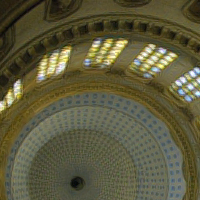}
        \caption{VGG-19 4th level \newline PSNR: 26.47dB}
        \label{exp1_G} 
                \end{subfigure}
                \hfill
                \begin{subfigure}[b]{0.24\textwidth}
        \includegraphics[width=\textwidth]{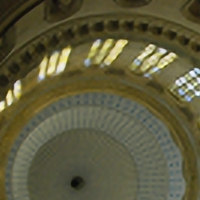}
        \caption{Baseline \newline PSNR: 26.91dB}
        \label{exp1_H}
                \end{subfigure}
\caption{Synthesized images. 
We compare the original images (first column) with sampled
synthesized images from Scattering (second column) and VGG (third column) networks as well
as the result of our baseline (fourth column) up-scaling $\times 3$, as a reference.}
\label{exp1}
\vspace{-3ex}
\end{figure}
\section{Experimental evaluation}
\label{experiments}

\subsection{Baseline and dataset}
\label{sec:setting}

As a baseline we use a CNN inspired in the one used by \cite{dong2014learning}. Specifically, a 4-layer CNN with \{64, 64, 64, 32\} feature maps and a linear output layer.
Filter sizes are $7\times 7$, $3\times 3$, $3\times 3$, and $5\times 5$ respectively, with ReLU non-linearities at each hidden layer. 
As a ``sanity-check'' we evaluated our trained architectures in the same benchmarks provided by \cite{dong2014learning} obtaining
comparable (slightly better) results in terms of PSNR.

All models were trained using $64\times 64$ images patches randomly chosen from a subset of the training set of ImageNet (\cite{deng2009imagenet}). Specifically we used 12.5M patches, extracting at most two patches per image. For testing we used images extracted from the test set of ImageNet and other external images.
 In this work we evaluate results using up-scaling factors of 3 and 4. To synthesize the low-resolution samples, the high-resolution images were sub-sampled using a proper anti-aliasing filters. 

\subsection{Perceptual relevance of the representation}

As discussed in Section~\ref{sec.stabel_inf}, there is a compromise between stability and discriminability in the design of the sufficient statistics. 
In this work, we argue that CNN are good candidates for this task. 
The only way to validate is to visually inspect what information the iso-probability sets are actually capturing about an image.
We do this by sampling different candidates, $r'$,  such that $\Psi(r')$ matches the ground truth features, $\Psi(r)$, for a high-resolution residual image $r$. 
Figure~\ref{exp1} shows the obtained results for two examples with very different image content. The image given in Figure~\ref{exp1_A} is a textured natural image
while Figure~\ref{exp1_B} has fine geometric patterns. 
As candidate representations we use Scattering CNN and the VGG network as discussed in sections~\ref{sec:scatt} and \ref{sec:vgg}. We initialize $r$ as Gaussian noise.
It can be observed that the synthesized images retain excellent perceptual quality.
As a reference, we included the results obtained when up-scaling these images (with factor of 3) using baseline CNN.
It is interesting to point out that the quality does not always correlate with MSE. In the example of Figure~\ref{exp1_A}, we see that the sampled image achieves smaller MSE producing a higher PSNR. For Figure~\ref{exp1_B} the PSNR of the sampled image is lower than both of the baselines, while captures much better the fine geometric structure. By inspecting carefully, one can see that the synthesized patterns in figures~\ref{exp1_F} and \ref{exp1_G} do not exactly match those in Figure~\ref{exp1_B} but their perceptual quality is much higher than the one of Figure~\ref{exp1_H}.

\begin{figure}
                \begin{subfigure}[b]{0.46\textwidth}
        \includegraphics[width=\textwidth]{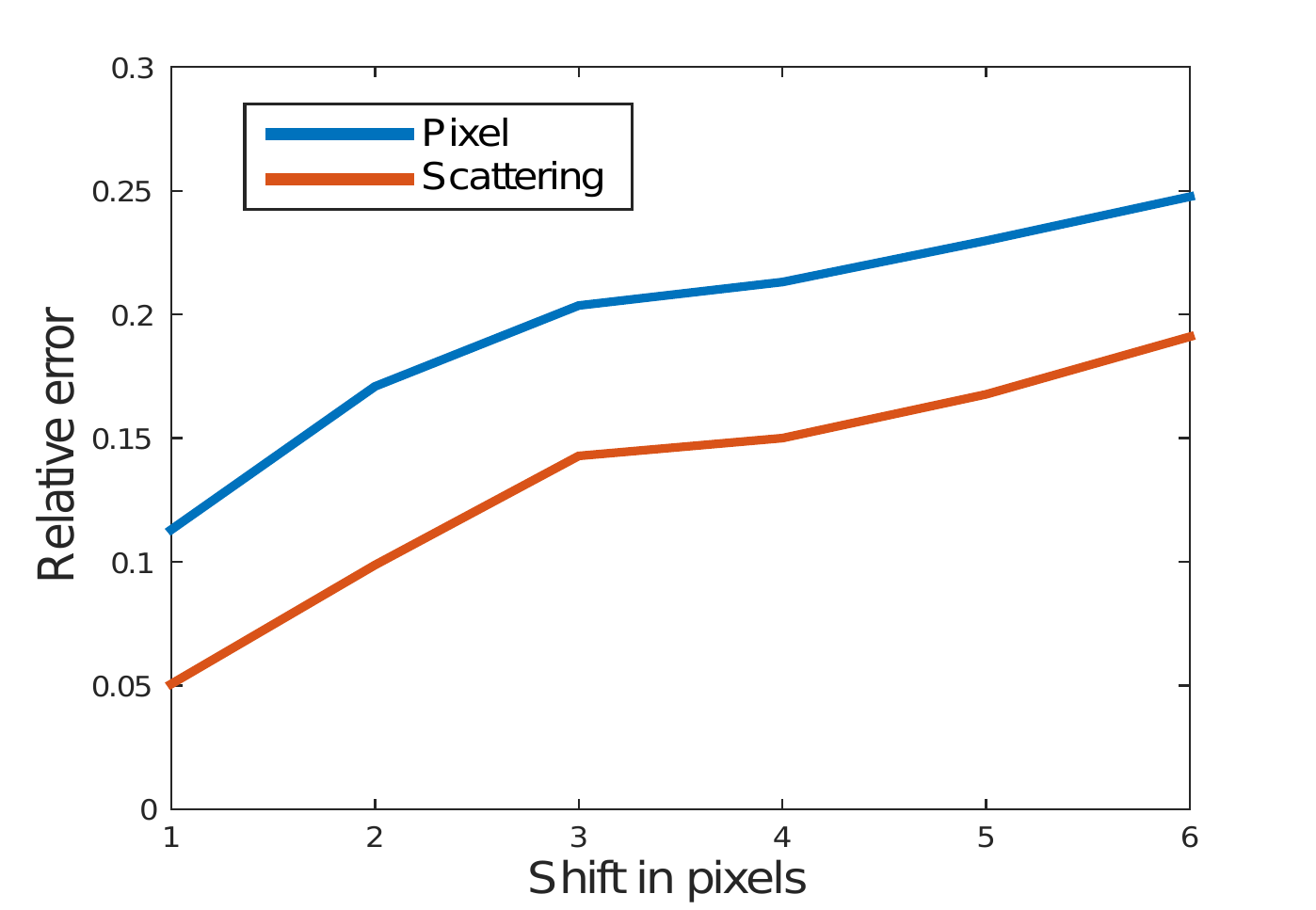}\\[-2ex]
                \end{subfigure}
        \hfill
        \begin{subfigure}[b]{0.46\textwidth}
\includegraphics[width=\textwidth]{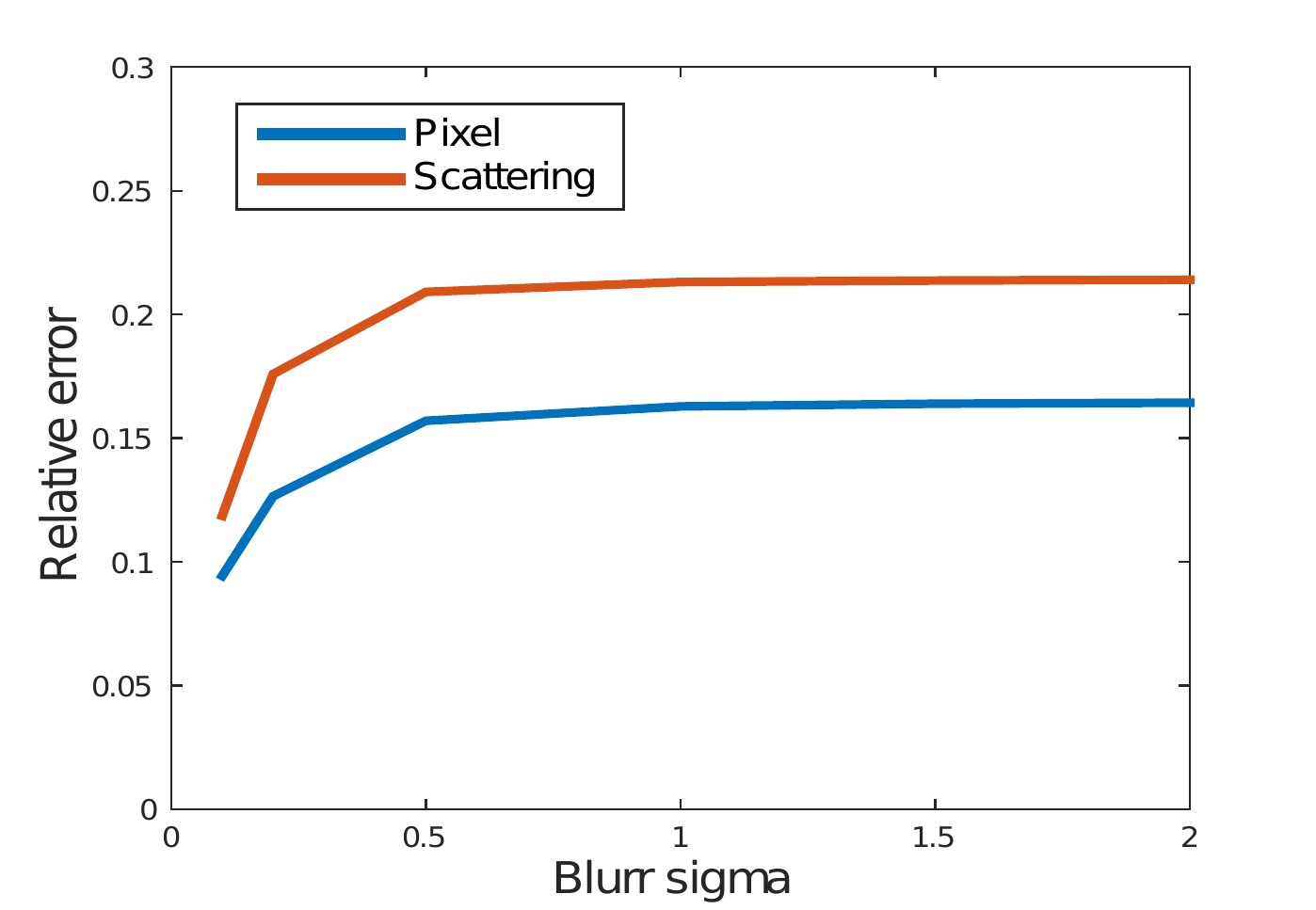}\\[-2ex]
        \end{subfigure}

\caption{Relative error in pixel and in scattering domains observed when applying image deformations to high-resolution images:
(left) rigid shift; (right) Gaussian blur. Relative error plotted agains the ``severity'' of the degradation measured in pixel shifts or standard deviation of the blur. 
Results are the average over 10 images of size $200\times 200$.}
\label{exp2}
\vspace{-2ex}
\end{figure}

The MSE is known to penalize heavily local image deformation while being more tolerant to blurring. To further stress this point we performed a simple experiment comparing
the relative distance of the residuales and their corresponding representations when undergoing degradations given by translation and blurring. 
Results are depicted in Figure~\ref{exp2}. We observe that both metrics present opposite behaviors. The MSE is less sensitive to blur and more sensitive to local translations.

\subsection{Super resolution}

The scattering network $\Psi$ is a 3-layer complex convolutional network that uses complex modulus as point-wise non-linearities. Its filters are near-analytic Morlet wavelets that span $8$ orientations and $3$ scales. We also include a feature map that captures the local total variation, as explained in Section \ref{model}. The resulting number of feature maps is $219$. We used a 5-layer CNN mimicking the scattering network for representing the function $\Phi$. 
We used an architecture with \{ 32, 64, 64, 64, 219\} feature maps with filter sizes $\{9\!\times\!9, 9\!\times\!9,9\!\times\!9,3\!\times\!3,1\!\times\!1\}$ respectively.
Unlike the scattering network, $\Phi$ has real-valued coefficients using ReLU non-linearity instead of the modulus operator.
Between the 2rd and 3rd layers, and between the 3rd and 4th layers, we included down-sampling layers. 
No explicit pooling layers were included in $\Phi$, even though the scattering CNN, $\Psi$, includes it its filter maps several low-pass and downsampling stages.
For the VGG-19 network, we use the same (given) architecture for both networks. 

\begin{figure}
\begin{subfigure}[b]{0.195\textwidth}
\includegraphics[width=\textwidth]{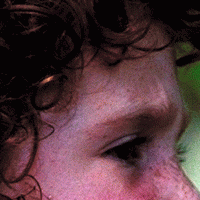}\\[1ex]
\includegraphics[width=\textwidth]{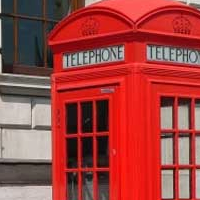}\\[1ex]
\includegraphics[width=\textwidth]{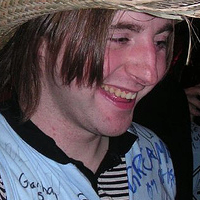}\\[1ex]
\includegraphics[width=\textwidth]{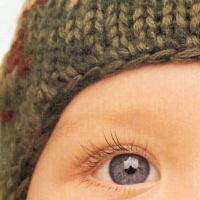}
\caption{Original}
\label{exp2_A}
\end{subfigure}
                \hfill
\begin{subfigure}[b]{0.195\textwidth}
\includegraphics[width=\textwidth]{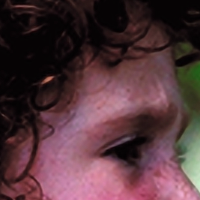}\\[1ex]
\includegraphics[width=\textwidth]{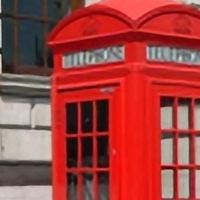}\\[1ex]
\includegraphics[width=\textwidth]{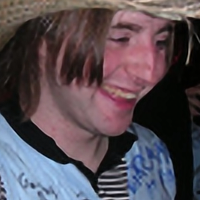}\\[1ex]
\includegraphics[width=\textwidth]{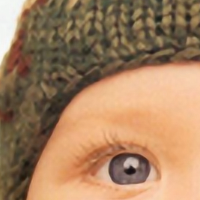}
\caption{Baseline}
\label{exp2_B}
\end{subfigure}
 \hfill
 \begin{subfigure}[b]{0.195\textwidth}
 \includegraphics[width=\textwidth]{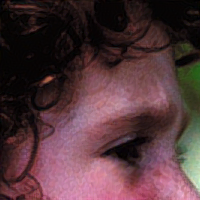}\\[1ex]
  \includegraphics[width=\textwidth]{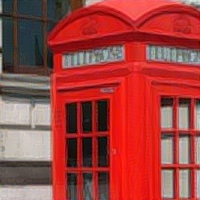}\\[1ex]
 \includegraphics[width=\textwidth]{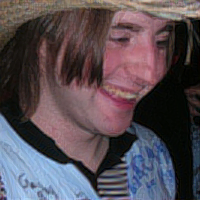}\\[1ex]
 \includegraphics[width=\textwidth]{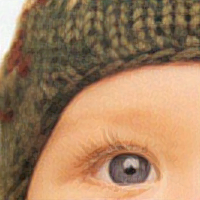}
 \caption{VGG-19}
 \label{exp2_C}
 \end{subfigure}
 \hfill
\begin{subfigure}[b]{0.195\textwidth}
\includegraphics[width=\textwidth]{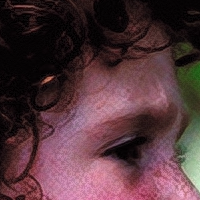}\\[1ex]
\includegraphics[width=\textwidth]{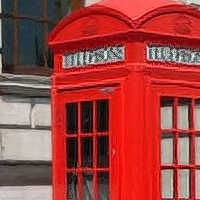}\\[1ex]
\includegraphics[width=\textwidth]{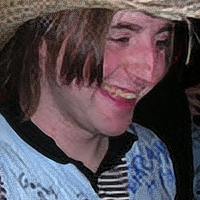}\\[1ex]
\includegraphics[width=\textwidth]{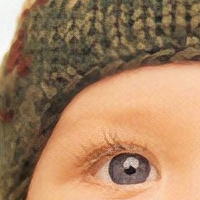}
\caption{Scattering}
\label{exp2_D}
\end{subfigure}
 \hfill
\begin{subfigure}[b]{0.195\textwidth}
\includegraphics[width=\textwidth]{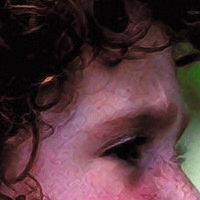}\\[1ex]
\includegraphics[width=\textwidth]{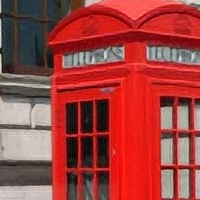}\\[1ex]
\includegraphics[width=\textwidth]{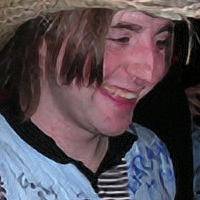}\\[1ex]
\includegraphics[width=\textwidth]{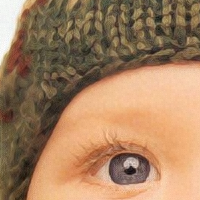}
\caption{Fine Tunned}
\label{exp2_E}
\end{subfigure}
\caption{Super-resolution results for up-scaling $\times 3$. 
We compare the original images (column (a) ), the baseline CNN (column (b) ) with sampled
synthesized images from our model using VGG (column (c) ), Scattering (column (d) ), and fine-funned Scattering (column (e) ) as CNNs.}
\label{exp3}
\vspace{-2ex}
\end{figure}
%
We first trained model given a fixed scattering and VGG networks. In this case, as discussed earlier, the problem can be approximated to minimizing the MSE as given in (\ref{bla2}).
Then ran a fine-tuning round for the scattering, as explained in Section~\ref{sec:fine}, by initializing the system using the pre-trained network as initial networks.
%
In the previous section we discussed the known problems of using the MSE as a way of measuring the perceptual relevance images.
Thus we limit our analysis to visual evaluation. The point estimate leads in all cases to better PSNR, which is to be expected as, unlike the proposed approach, the model
is optimized for maximizing this quantity.
Figure~\ref{exp3} depicts exemplar results for super-resolution problem for an up-scaling factor of $\times 3$ ( results for up-scaling $\times 4$ are provided in Appendix~\ref{experiments_appendix}), which represents a good benchmark for inferential models, since the point estimates start to exhibit serious regression to the mean.
We compare examples synthesized from predicted Scattering and VGG networks, a fine-funned scattering network, as well as our baseline CNN.
It is visually apparent that the predictions by the proposed model are able to produce results with more stable high-frequency content, avoiding the ``regression to the mean'', observable in the baseline results. The fine-tuned version of our model is able to reduce artifacts present in the original version, this is further analyzed in Appendix~\ref{experiments_appendix}. This effect
becomes even more clear when comparing the residual images (with respect to the bicubic up-sampling), as shown in Appendix~\ref{experiments_appendix}.
The proposed method, has some natural limitations. When tested with very fine textures, the recovered high frequencies appear artificial, see Appendix~\ref{experiments_appendix} for some examples. Finally, from a computational perspective, the baseline CNN is more efficient than the proposed approach. The proposed approach requires solving an inference problem as explained in Section~\ref{sec:sampling}. To give a practical idea, upscaling a factor of 3 an image of size $200\times 200$ takes 0.1 seconds for the baseline CNN, 5.26 for the Scattering and 5.09 seconds for the VGG. Reported times are average over fifty runs and were run on a  Nvidia GTX Titan Black GPU, using 100 iterations of gradient decent.


\section{Discussion}
\label{conclusions}

Generating realistic high-frequency content is a hard problem due to the curse of dimensionality. In this paper, we have argued that sharp geometric structures and textured regions are not well approximated by low-dimensionality models -- if that was the case, point estimates trained on low-resolution patches would scale better. 
%
In order to overcome this problem, we have proposed a conditionally generative model that uses sufficient statistics from CNNs  to characterize with stable features textures and high-frequency content. 
 By properly initializing the networks with filters with good geometrical properties, such as in scattering networks, we obtain a good baseline that generates realistic high frequency content. The underlying uncertainty in the generation of output samples is in our case encoded in the multiple combinations of complex phases in the intermediate layers of the network that are compatible with the output features. Recent work on signal recovery from these non-linear representations (\cite{pooling_recovery}) suggests that the effective dimensionality of this set of admissible phases ranges from low (when the redundancy of the network is high) to nearly the ambient dimension (when the redundancy is low), thus suggesting that these models could scale well.


We proposed an algorithm to fine-tune those sufficient statistics to the data, by optimizing a surrogate for the conditional likelihood. Although the model appears to move in the right direction, as shown in the numerical experiments, there is still a lot to be done and understood. The fine-tuning step is costly and we observed the learning parameters (such as the learning rates of each network) need to be adjusted carefully, due mostly to the bias and variance of the gradient estimates. Another valid critique of the model is that the test-time inference is more expensive than a point estimate, or a generative adversarial network alternative. 
The upside is that the inference enforces solutions with spatial coherence, and it is not obvious that a feedforward procedure is able to align high frequency content without any feedback loop. Also, the proposed model provides an explicit representation.

This paper focused on super-resolution, but the underlying challenge is how to find an appropriate metric that is compatible with the high-frequency content of natural images. This poses a fundamental trade-off between sharpness and stability: the phase of high-frequency coefficients (encoding the precise location of sharp structures) is fundamentally unstable to small deformations and thus unpredictable; therefore, restoring high-frequency information comes at the cost of misalignments.

\bibliography{iclr2016_conference}
\bibliographystyle{iclr2016_conference}

\begin{appendices} 

\section{Further experimental evaluation}
\label{experiments_appendix}
\begin{figure}
\begin{subfigure}[h!]{0.32\textwidth}
\includegraphics[width=\textwidth]{figs/exp3_file3_GT.PNG}
\caption{Original}
\end{subfigure}
                \hfill
\begin{subfigure}[h!]{0.32\textwidth}
\includegraphics[width=\textwidth]{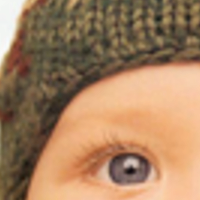}
\caption{Bicubic}
\end{subfigure}
 \hfill
 \begin{subfigure}[h!]{0.32\textwidth}
\includegraphics[width=\textwidth]{figs/exp3_file3_PE.PNG}
\caption{Baseline}
 \end{subfigure}\\[0.4ex]
\begin{subfigure}[h!]{0.32\textwidth}
 \includegraphics[width=\textwidth]{figs/exp3_file3_scatt.PNG}
\caption{Scattering}
\end{subfigure}
                \hfill
\begin{subfigure}[h!]{0.32\textwidth}
 \includegraphics[width=\textwidth]{figs/exp3_file3_scatt_fine.PNG}
\caption{Scattering fine-tunned}
\end{subfigure}
 \hfill
 \begin{subfigure}[h!]{0.32\textwidth}
   \includegraphics[width=\textwidth]{figs/exp3_file3_vgg_average.PNG}
\caption{VGG-19}
 \end{subfigure}
\caption{Synthesis results sale $\times 3$. Images $200\times 200$ pixels. Residual images provided in Figure~\ref{appA2}}
\label{appA1}
\vspace{-3ex}
\end{figure}

\begin{figure}
\begin{subfigure}[h!]{0.32\textwidth}
\centering
\vspace{0.335\textwidth}
\includegraphics[width=0.33\textwidth]{figs/exp3_file3_LR.PNG}
\vspace{0.335\textwidth}
\caption{Original size}
\end{subfigure}
                \hfill
\begin{subfigure}[h!]{0.32\textwidth}
\includegraphics[width=\textwidth]{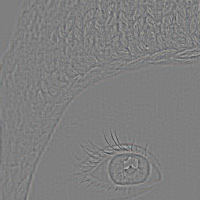}
\caption{Ground truth}
\end{subfigure}
 \hfill
 \begin{subfigure}[h!]{0.32\textwidth}
\includegraphics[width=\textwidth]{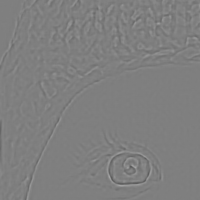}
\caption{Baseline}
 \end{subfigure}\\[0.4ex]
\begin{subfigure}[h!]{0.32\textwidth}
 \includegraphics[width=\textwidth]{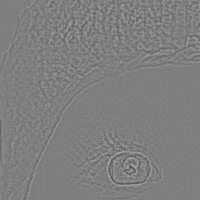}
\caption{Scattering}
\end{subfigure}
                \hfill
\begin{subfigure}[h!]{0.32\textwidth}
 \includegraphics[width=\textwidth]{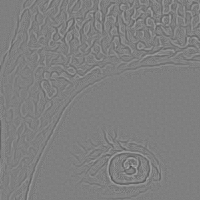}
\caption{Scattering fine-tunned}
\label{res_fine}
\end{subfigure}
 \hfill
 \begin{subfigure}[h!]{0.32\textwidth}
   \includegraphics[width=\textwidth]{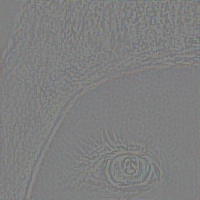}
\caption{VGG-19}
 \end{subfigure}
\caption{Residual image with respect to the Bicubic interpolation, corresponding to Figure~\ref{appA1}. Top left original sized image.}
\label{appA1res}
\vspace{-3ex}
\end{figure}

\begin{figure}
\begin{subfigure}[h!]{0.32\textwidth}
\includegraphics[width=\textwidth]{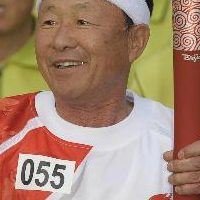}
\caption{Original}
\end{subfigure}
                \hfill
\begin{subfigure}[h!]{0.32\textwidth}
\includegraphics[width=\textwidth]{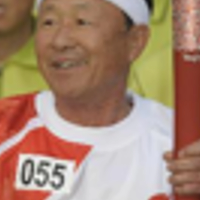}
\caption{Bicubic}
\end{subfigure}
 \hfill
 \begin{subfigure}[h!]{0.32\textwidth}
\includegraphics[width=\textwidth]{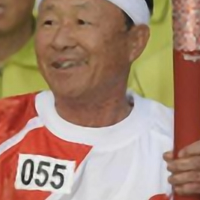}
\caption{Baseline}
 \end{subfigure}\\[0.4ex]
\begin{subfigure}[h!]{0.32\textwidth}
 \includegraphics[width=\textwidth]{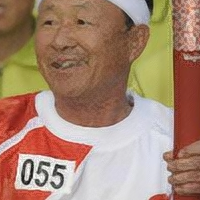}
\caption{Scattering}
\end{subfigure}
                \hfill
\begin{subfigure}[h!]{0.32\textwidth}
 \includegraphics[width=\textwidth]{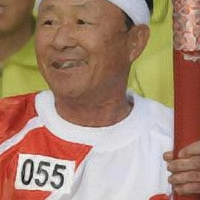}
\caption{Scattering fine-tunned}
\end{subfigure}
 \hfill
 \begin{subfigure}[h!]{0.32\textwidth}
   \includegraphics[width=\textwidth]{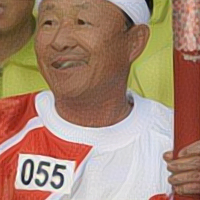}
\caption{VGG-19}
 \end{subfigure}
\caption{Synthesis results sale $\times 3$.}
\label{appA2}
\vspace{-3ex}
\end{figure}

\begin{figure}
\begin{subfigure}[h!]{0.32\textwidth}
\includegraphics[width=\textwidth]{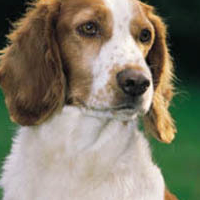}
\caption{Original}
\end{subfigure}
                \hfill
\begin{subfigure}[h!]{0.32\textwidth}
\includegraphics[width=\textwidth]{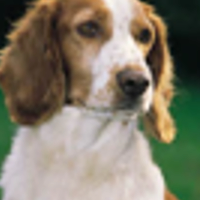}
\caption{Bicubic}
\end{subfigure}
 \hfill
 \begin{subfigure}[h!]{0.32\textwidth}
\includegraphics[width=\textwidth]{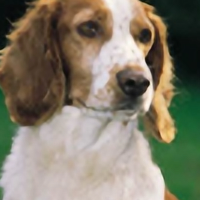}
\caption{Baseline}
 \end{subfigure}\\[0.4ex]
\begin{subfigure}[h!]{0.32\textwidth}
 \includegraphics[width=\textwidth]{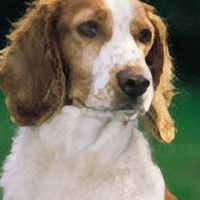}
\caption{Scattering}
\end{subfigure}
                \hfill
\begin{subfigure}[h!]{0.32\textwidth}
 \includegraphics[width=\textwidth]{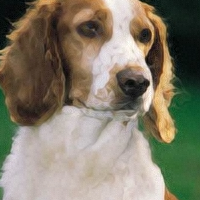}
\caption{Scattering fine-tunned}
\end{subfigure}
 \hfill
 \begin{subfigure}[h!]{0.32\textwidth}
   \includegraphics[width=\textwidth]{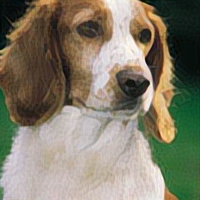}
\caption{VGG-19}
 \end{subfigure}
\caption{Synthesis results sale $\times 3$.}
\label{optional_1}
\vspace{-3ex}
\end{figure}

\begin{figure}
\begin{subfigure}[h!]{0.32\textwidth}
\includegraphics[width=\textwidth]{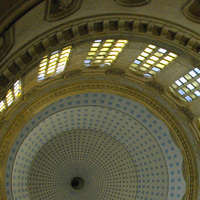}
\caption{Original}
\end{subfigure}
                \hfill
\begin{subfigure}[h!]{0.32\textwidth}
\includegraphics[width=\textwidth]{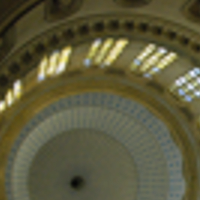}
\caption{Bicubic}
\end{subfigure}
 \hfill
 \begin{subfigure}[h!]{0.32\textwidth}
\includegraphics[width=\textwidth]{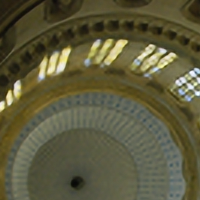}
\caption{Baseline}
 \end{subfigure}\\[0.4ex]
\begin{subfigure}[h!]{0.32\textwidth}
 \includegraphics[width=\textwidth]{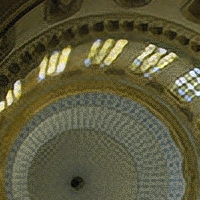}
\caption{Scattering}
\end{subfigure}
                \hfill
\begin{subfigure}[h!]{0.32\textwidth}
 \includegraphics[width=\textwidth]{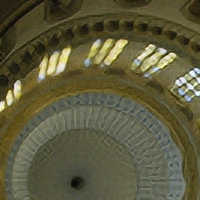}
\caption{Scattering fine-tunned}
\end{subfigure}
 \hfill
 \begin{subfigure}[h!]{0.32\textwidth}
   \includegraphics[width=\textwidth]{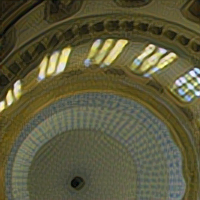}
\caption{VGG-19}
 \end{subfigure}
\caption{Synthesis results sale $\times 3$. Challenging example with fine texture.}
\label{texture}
\vspace{-3ex}
\end{figure}

\begin{figure}
\begin{subfigure}[h!]{0.32\textwidth}
\includegraphics[width=\textwidth]{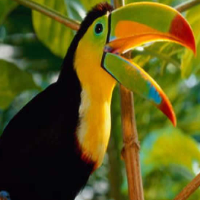}
\caption{Original}
\end{subfigure}
                \hfill
\begin{subfigure}[h!]{0.32\textwidth}
\includegraphics[width=\textwidth]{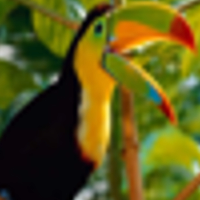}
\caption{Bicubic}
\end{subfigure}
 \hfill
 \begin{subfigure}[h!]{0.32\textwidth}
\includegraphics[width=\textwidth]{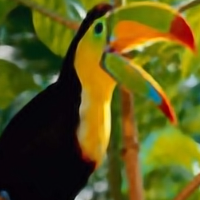}
\caption{Baseline}
 \end{subfigure}\\[0.4ex]
\begin{subfigure}[h!]{0.32\textwidth}
 \includegraphics[width=\textwidth]{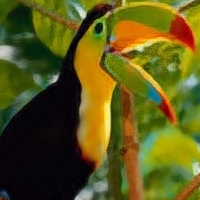}
\caption{Scattering}
\end{subfigure}
                \hfill
\begin{subfigure}[h!]{0.32\textwidth}
 \includegraphics[width=\textwidth]{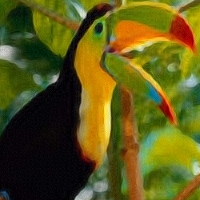}
\caption{Scattering fine-tunned}
\end{subfigure}
 \hfill
 \begin{subfigure}[h!]{0.32\textwidth}
   \includegraphics[width=\textwidth]{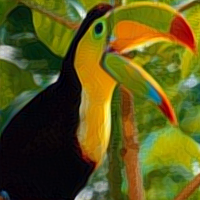}
\caption{VGG-19}
 \end{subfigure}
\caption{Synthesis results sale $\times 4$.}
\label{times4}
\vspace{-3ex}
\end{figure}

\begin{figure}
\begin{subfigure}[h!]{0.32\textwidth}
\includegraphics[width=\textwidth]{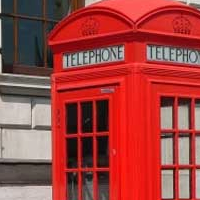}
\caption{Original}
\end{subfigure}
                \hfill
\begin{subfigure}[h!]{0.32\textwidth}
\includegraphics[width=\textwidth]{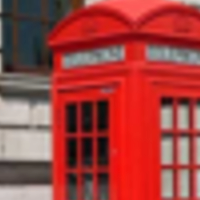}
\caption{Bicubic}
\end{subfigure}
 \hfill
 \begin{subfigure}[h!]{0.32\textwidth}
\includegraphics[width=\textwidth]{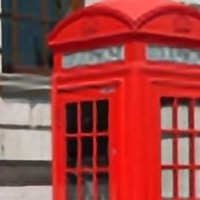}
\caption{Baseline}
 \end{subfigure}\\[0.4ex]
\begin{subfigure}[h!]{0.32\textwidth}
 \includegraphics[width=\textwidth]{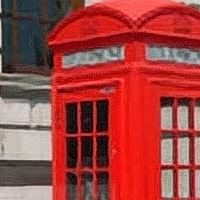}
\caption{Scattering}
\end{subfigure}
                \hfill
\begin{subfigure}[h!]{0.32\textwidth}
 \includegraphics[width=\textwidth]{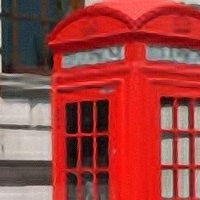}
\caption{Scattering fine-tunned}
\end{subfigure}
 \hfill
 \begin{subfigure}[h!]{0.32\textwidth}
   \includegraphics[width=\textwidth]{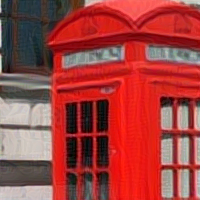}
\caption{VGG-19}
 \end{subfigure}
\caption{Synthesis results sale $\times 4$.}
\label{times4_bis}
\vspace{-3ex}
\end{figure}

In this appendix we provide further experimental evaluation and analysis that could not be included in the main body of the paper due to space limitations.
Figures~\ref{appA1}-\ref{texture} show exemplar results for up-scaling $\times 3$, while figures~\ref{times4} and \ref{times4_bis} provide results for $\times 4$ magnification. Different networks were trained for different up-scaling factors. We compare the same models as in Section~\ref{experiments}.

Figure~\ref{appA1} is a detailed version the bottom image in Figure~\ref{exp3}. This image has been widely used as a benchmark for the super-resolution problem.
It is interesting because it has sharp edges, fine details and textures. The proposed approach is able to synthesize more high-frequency content than the baseline approach.
This can be well observed in the texture in the hat as well as the eye. 
This can be also seen in the details of the torch in Figure~\ref{appA2}, which are completely lost in image recovered by the baseline CNN. 
A natural risk for the proposed approach is to produce results with high frequency content that looks un-natural. Both Scattering and VGG networks can suffer from this problem. We observe that this effect is a bit stronger in the latter. This can be appreciated in figures~\ref{appA2} and~\ref{optional_1}.
The fine-tuning stage is able to significantly reduce  this effect. To better show this feature, Figure~\ref{appA1res} shows the high frequency content added by each method. The residual in Figure~\ref{res_fine} appears smoother while maintaining the sharpness and fine texture. 

We included two examples as a way of showing images in which adding structure can lead to perceptually noticeable artifacts. The first one is the very challenging example first discussed
in Figure~\ref{exp1}, here shown in Figure~\ref{texture}. This image has very fine textures, whose information is almost completely lost in the low resolution image. 
The proposed method for both, VGG and Scattering networks, is able to predict the presence of high-frequency content, however the results exhibit noticeable artifacts.
 Interestingly the fine tuning seems to hurt despite sharpening the image. This example is particularly challenging as the texture is highly structured and human observers expect to see a very defined pattern. The second example shows an image where the high-resolution version contains blurred areas (due to defocus of the background). While producing
a good reconstruction of the bird, the proposed approach tends to over sharpen these regions, as there is no way to know a-priori where (and where not) to sharpen. Arguably, producing a blurred image can be a ``better'' alternative.

\newpage

\newpage

%

\newpage

\clearpage

\section{Training Details}
In this Appendix we provide further details of our training implementation. 
\begin{itemize}
\item Fine-tuning: We alternate between optimizing $\Phi$ and $\Psi$. We adjust the learning rate corresponding to the parameters of $\Psi$ by a factor $\eta = 10^{-4}$. 
\item We use Adam (\cite{adam}) to perform the inference to generate negative samples. 
\item We adjust the shrinkage of the total variation features at test time by adding a term $\lambda \| r \|_{\textrm{TV}}$, with $\lambda = 10^{-8}$.
\item We renormalize the scattering features by scaling each output feature by $c^k$, where $c>1$ and 
$k$ is the number of nonlinearities corresponding to each scattering path ($k=0,1, 2$ in our experiments). 
\end{itemize}

%
%

\end{appendices}
\end{document}